\documentclass[letterpaper, 10 pt, journal, twoside]{IEEEtran}
\ifCLASSINFOpdf
\else
\fi
\usepackage{float}
\usepackage{amsfonts}
\usepackage{amsthm}
\usepackage{pifont}
\usepackage{dblfloatfix}
\usepackage{xcolor}
\usepackage{soul}

\newtheorem{definition}{Definition}

\newtheorem*{problem*}{Problem}
\usepackage[colorinlistoftodos]{todonotes}
\usepackage{glossaries}
\usepackage{listings}
\usepackage{amsmath}
\usepackage{amssymb}
\usepackage[linesnumbered,ruled]{algorithm2e}
\usepackage{graphicx}
\usepackage{stix}

\usepackage{tikz}
\usepackage{etoolbox}
\usetikzlibrary{shapes,arrows}
\usepackage{verbatim}
\usepackage[hidelinks]{hyperref}

\newcommand{\deleted}[1]{}

\definecolor{SpringGreen}{RGB}{40,160,80} % Define SpringGreen
\definecolor{SkyBlue}{RGB}{60,130,200}   % Define SkyBlue
\definecolor{Khaki}{RGB}{160,100,70}     % Define Khaki
\definecolor{Lavender}{RGB}{180,100,180}  % Define Lavender
\newcommand{\replytoall}[1]{\textcolor{SkyBlue}{#1}}
\newcommand{\replytofour}[1]{\textcolor{SpringGreen}{#1}}
\newcommand{\replytofive}[1]{\textcolor{Khaki}{#1}}
\newcommand{\replytoseven}[1]{\textcolor{Lavender}{#1}}
\renewcommand{\replytoall}[1]{{#1}}
\renewcommand{\replytofour}[1]{{#1}}
\renewcommand{\replytofive}[1]{{#1}}
\renewcommand{\replytoseven}[1]{{#1}}

\begin{document}
\title{AirBender: Adaptive Transportation of Bendable Objects Using Dual UAVs}

% \author{Michael~Shell,~\IEEEmembership{Member,~IEEE,}
%         John~Doe,~\IEEEmembership{Fellow,~OSA,}
%         and~Jane~Doe,~\IEEEmembership{Life~Fellow,~IEEE}% <-this % stops a space
% \thanks{M. Shell was with the Department
% of Electrical and Computer Engineering, Georgia Institute of Technology, Atlanta,
% GA, 30332 USA e-mail: (see http://www.michaelshell.org/contact.html).}% <-this % stops a space
% \thanks{J. Doe and J. Doe are with Anonymous University.}% <-this % stops a space
% \thanks{Manuscript received April 19, 2005; revised August 26, 2015.}}
\author{Jiawei Xu, Longsen Gao, Rafael Fierro, and David Salda\~{n}a
\thanks{Manuscript received: September, 15, 2024; Revised December, 9, 2024; Accepted January, 13, 2025.}%Use only for final RAL version
\thanks{This paper was recommended for publication by Editor Giuseppe Loianno upon evaluation of the Associate Editor and Reviewers' comments.} %Use only for final RAL version
\thanks{J. Xu and D. Salda\~{n}a are with the Autonomous and Intelligent Robotics Laboratory (AIRLab), Lehigh University, PA, 18015, USA:\texttt{\{jix519, saldana\}@lehigh.edu}, L. Gao and R. Fierro are with the Electrical and Computer Engineering Department, The University of New Mexico, Albuquerque, NM, 87131, USA:\texttt{\{lgao1, rfierro\}@unm.edu}.
\newline
J. Xu, and D. Saldaña gratefully acknowledge the support of the NSF Award 2322840.
\indent L. Gao and R. Fierro were supported by the Air Force Research Laboratory (AFRL) under agreements FA9453-18-2-0022 and FA9550-22-1-0093. Any opinions findings, and conclusions or recommendations expressed in this material are those of the authors and do not necessarily reflect the views of the United States Air Force.
}
\thanks{Digital Object Identifier (DOI): see top of this page.}
}

\markboth{IEEE Robotics and Automation Letters. Preprint Version. Accepted January, 2025}
{Xu \MakeLowercase{\textit{et al.}}: AirBender: Adaptive Transportation of Bendable Objects Using Dual UAVs} 




\maketitle
\begin{abstract}
The interaction of robots with bendable objects in midair presents significant challenges in control, often resulting in performance degradation and potential crashes, especially for aerial robots due to their limited actuation capabilities and constant need to remain airborne. This paper presents an adaptive controller that enables two aerial vehicles to collaboratively follow a trajectory while transporting a bendable object without relying on explicit elasticity models. %Our approach circumvents the performance limitations of traditional rigid-body or finite-element models and the lack of versatility of explicit elastic models. 
Our method allows on-the-fly adaptation to the object's unknown deformable properties, ensuring stability and performance in trajectory-tracking tasks. We use Lyapunov analysis to demonstrate that our adaptive controller is asymptotically stable.
Our method is evaluated through hardware experiments in various scenarios, demonstrating the capabilities of using multirotor aerial vehicles to handle bendable objects.
\end{abstract}


% \fi
\IEEEpeerreviewmaketitle

\section{Introduction}
    \IEEEPARstart{I}n recent years, multirotor aerial vehicles have rapidly advanced, finding applications in diverse fields such as environmental monitoring, surveillance,
    % ~\cite{boon2017comparison}
    package delivery,
    % ~\cite{saunders2024autonomous}
    and search-and-rescue operations.
    % ~\cite{erdos2013experimental}
    While conventional studies have focused mainly on rigid bodies and point-mass models for multirotor physical interactions~\cite{6875943,7989608}, real-world scenarios often involve nonrigidity, introducing deformations and elastic properties that significantly influence interaction outcomes~\cite{arriola2020modeling,yin2021modeling}. 
    As manipulating bendable objects requires considering deformations and variability in material properties, there are novel potential applications for robots to explore. For example, in agriculture, the deformation of tree branches poses challenges for autonomous cultivation~\cite{10380659}. In autonomous construction, the ability to manipulate bendable materials expands the variety of tasks robots can handle~\cite{radoglou2020compilation,6225012}. Our vision is to integrate bendable objects into aerial tasks, enhancing the versatility of aerial vehicles in physical interactions.

    Elastic deformation have been studied for almost four centuries~\cite{euler1980rational}, and still, they present nontrivial challenges to be incorporated into robotic systems, such as: \textit{i)} Nonlinear deformations are difficult to model and predict~\cite{Levien2008TheEA,KirchhoffRods}; 
    \textit{ii)} real-time sensing and feedback often require accurate detection of the shape and its deformation; 
    \textit{iii)} control algorithms must adapt to the unknown behavior of deformable objects;
    \textit{iv)} The variability in the physical properties of deformable objects, such as elasticity and stiffness, can change during manipulation. 
    %Addressing non-rigidity introduces \replytofive{additional} non-linear \replytofive{partial} differential equations into the dynamics of robotic systems~, resulting in performance degradation when using traditional control techniques. 
    Researchers have made advances to handle these challenges, including the estimation of elastic shapes~\cite{8206139,8593946} and the adept manipulation of bendable rods with robotic arms~\cite{bretl2014rod}. Model-based analyses often require additional parameters to describe nonrigidity. For example, Kirchhoff and Cosserat deformation models~\cite{8206139} offer generalizability but pose additional analytical challenges compared to the linear elastic rod model~\cite{6385876,9888782}. As alternatives, adaptive control~\cite{aghajanzadeh2022adaptive} and learning~\cite{9812244,8584049} make generic assumptions about the characteristics of bendable objects and demonstrate versatility and effectiveness in practical applications.

    \begin{figure}
        \centering
        \includegraphics[width=0.95\linewidth]{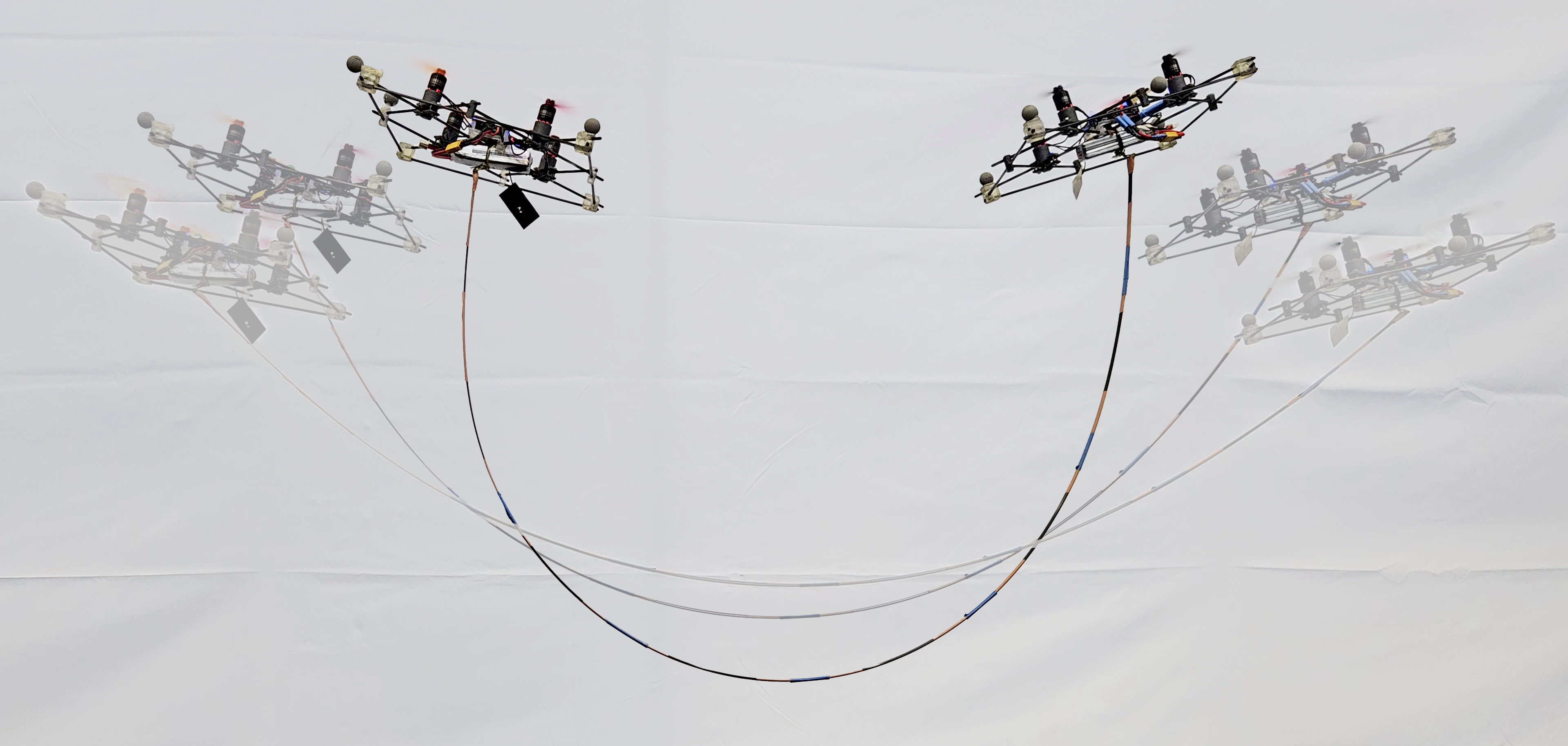}
        \caption{Two quadrotors actively control the endpoint trajectories of a long carbon-fiber strip using the proposed adaptive controller. \\
        \textbf{Video: \url{https://www.youtube.com/watch?v=lhTVIWSjhvQ}}}
        \label{fig:main}
    \end{figure}
    % On top of handling deformable objects with robots, using aerial robots adds another layer of intricacy~\cite{9955376} as well as . 
    For transportation purposes, the ability to manipulate non-rigid objects in midair significantly expands the range of objects and applications.
    % One would expect higher flexibility using aerial vehicles in transporting and manipulating bendable objects compared to fixed station robots due to their unobstructed access to $3$-D space. 
    However, the natural instability of aerial vehicles requires active stabilization efforts~\cite{1302409}. The presence of external disturbances, such as the elastic force of a bendable rod, can lead to manipulation failures or vehicle crashes. Additionally, the computational overhead of intense numerical methods may cause aerial vehicles to destabilize due to the inherent latency sensitivity of their motion control~\cite{7574310}. As a result, few attempts have been made to enable aerial vehicles to manipulate bendable objects in midair, mostly using explicit models of specific elastic objects, both theoretical~\cite{goodman2022geometric} and practical~\cite{G005914}.
%
    % Recent advances in the literature have highlighted the effectiveness of deploying adaptive controllers on aerial vehicles to counter unknown disturbances in practical settings~\cite{9571068,chang2023new,9138676}. The major advantages of adaptive controllers for aerial vehicles are the low computational overhead and the performance guarantee~\cite{khalil2014nonlinear}. In addition, their robustness allows aerial vehicles to adapt to intricate external disturbances using simpler reference models~\cite{estevez2017online}. 
    We propose an adaptive control approach that allows two aerial vehicles to collaboratively transport a bendable strip in real-time. 
    The object's deformable properties are unknown, and robots are tasked to follow a trajectory while holding the strip from its endpoints. 
    %More specifically, given a bendable object with unknown physical characteristics other than its tendency to rebound to its natural shape, we connect a multirotor vehicle at each of the rod's endpoints and follow a desired trajectory.
    %Such an action resembles bending construction materials to pass narrow space or tree branches to reachable working areas of a cultivation machine.  
    
    \replytoall{
    The main contribution of this paper is the development of a stable trajectory-tracking controller that leverages the recursive least square (RLS) approximation to adapt to an unknown external force from deformation. 
    %Specifically, we design the feature vectors of RLS for the scenario where two multirotor aerial vehicles control the trajectory of the endpoints of a bendable object.
    }    
    \replytofour{
    In comparison to works in multi-robot transportation of rigid payloads~\cite{loianno2017cooperative,https://doi.org/10.1155/2022/2486561,0278364919854131,8294249}, our proposed method considers the deformation of the payload and the unknown induced elastic force. %In addition, we consider that the connection between the bendable object and the aerial vehicle constrains only the position, inducing only a force, but not a torque to adapt to.     
    Our controller operates without requiring an explicit elasticity model, continuous~\cite{6696850,10615146} or discretized~\cite{9517345,G005914} and adapts to unknown mass, density, and Young's modulus. 
    } 

\section{Problem statement} 
    \label{sec:p}
    The goal of this work is to transport a bendable object, e.g., a rod, or strip, using two aerial vehicles.
    % The object is bendable and its material properties are unknown.
    
    \begin{definition}[vehicle]
        A vehicle is a multirotor with a rigid frame, total weight $m$, and inertia tensor $\boldsymbol{J}$. 
        Its rotors generate thrust to translate and rotate the vehicle.
    \end{definition}
    \noindent
    The vehicles are located at the ends of the object. We model the connection between each vehicle and the object as a passive spherical joint. 
 % It is important to distinguish this constraint from the one typically seen in a cantilever model, as discussed in other research (see~\cite{9955376,aghajanzadeh2022adaptive}), where fixing one end of the object in position and orientation indicates both a torque and a force exerted on the object.
    \begin{definition}[Bendable object]
        A bendable object is a long, thin, \replytofive{unstretchable, and incompressible} elastic object with a curve length $L_0$. 
        %Its thickness $0 < w \ll L_0$ is negligible compared to its curve length.
        Its thickness $w$, mass, $m_b$, and mass distribution are unknown.
        The elasticity properties, such as Young's modulus, are also unknown.       
    \end{definition}
    %%%%%%%%%%%%%%%%%%%%%%%%%%%%%%%%%%%%%%%%%%%% 
    \begin{figure}[t]
        \centering
        {\includegraphics[width=\linewidth]{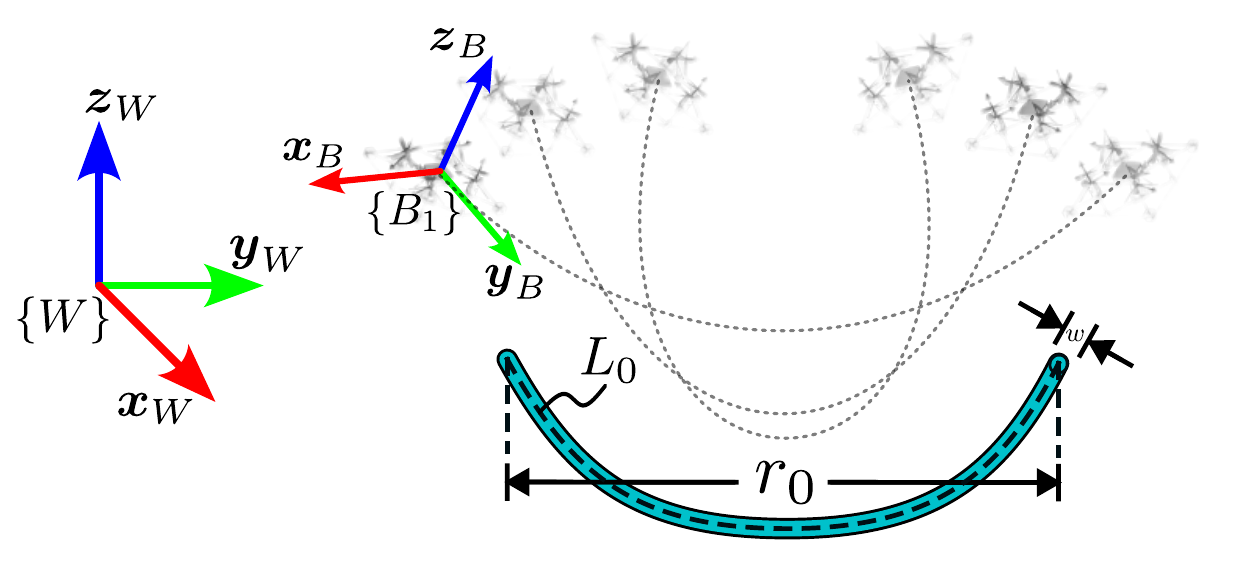}}
        \caption{The vehicles apply forces on two ends of an object, resulting in different bending curves. We consider the world reference frame $\{W\}$ and the body frame of each vehicle, $\{B_1\}$ and $\{B_2\}$.
        } 
        \label{fig:bendable}
    \end{figure}
    %%%%%%%%%%%%%%%%%%%%%%%%%%%%%%%%%%%%%%%%%%%%
    %A multirotor is capable of flying independently. In addition to controlling its own motion, a multirotor can apply force and torque to a payload when connected to it.
    \noindent
    Considering the two vehicles connected to the two endpoints of a bendable object, 
    we index the two endpoints and the corresponding vehicle with $i = 1, 2$. 
    The body frame $\{B_i\}$ of the vehicle $i$ has its origin fixed at its center of mass (COM). 
    The $z$-axis of $\{B_i\}$ points in the upward direction of the vehicle, and the $x$-axis points in the ``front'' direction of a vehicle. 
    The vehicle generates a force $\boldsymbol{f}_i$ and a torque~$\boldsymbol{\tau}_i$.
    
    The position of the $i$-th vehicle in $\{W\}$ is denoted by $\boldsymbol{p}_i$, and the orientation of $\{B_i\}$ with respect to $\{W\}$ is denoted by the rotation matrix $\boldsymbol{R}_i\in\mathsf{SO(3)}$.
   
    We assume that each endpoint is directly fixed in the COM of a vehicle, resulting in the \emph{endpoint displacement vector}, $\boldsymbol{r} = \boldsymbol{p}_2 - \boldsymbol{p}_1$. 
    The magnitude of the endpoint displacement is denoted by $r = \Vert\boldsymbol{r}\Vert$, satisfying $0\leq r\leq L_0$.
    The vehicles are affected by the object's force $\boldsymbol{f}_i^o$ and torque $\boldsymbol{\tau}_i^o$.
    Since the object is connected to the COM, the object generates zero torque on the vehicle, \textit{i.e.}, $\boldsymbol{\tau}_i^o = \boldsymbol{0}$.
    
    The object has a natural distance between two endpoints, denoted by $r_0$, when an external force other than gravity is applied. \replytofive{In this state, $r_0 \leq L_0$ due to the curvature introduced by the bending. When external forces are applied to the endpoints, the object exerts an elastic force as it tends to return to its natural endpoint distance $r_0$. These characteristics are unknown to the vehicles in our setup, as no explicit elasticity model or knowledge of $L_0$ or $r_0$ is assumed.} 
    %from bending the object without knowing its physical characteristics other than that it is bendable and non-extendable beyond its curve length, including its mass distribution and elasticity. 
    
    We describe the vehicle dynamics using Newton-Euler equations,
    \begin{eqnarray}
        m\ddot{\boldsymbol{p}_i} + mg\boldsymbol{z} &=& \boldsymbol{f}_i + \boldsymbol{f}_i^o,\label{eq:newton}\\
        \boldsymbol{J}\dot{\boldsymbol{\omega}_i} + \boldsymbol{\omega}_i\times\boldsymbol{J\omega}_i &=& \boldsymbol{\tau}_i, \label{eq:euler}
    \end{eqnarray}
    where $\boldsymbol{\dot{p}}_i, \boldsymbol{\Ddot{p}}_i$ are the linear velocity and acceleration; $\boldsymbol\omega_i, \boldsymbol{\dot{\omega}_i}$ are angular velocity and acceleration, respectively. The term~$g$ is the gravitational constant.
    % Since we consider the linear motion in the world reference frame and the rotation in the multirotor frame, $\boldsymbol{f}_i = \boldsymbol{R}_i{}^B\boldsymbol{f}_i$, $\boldsymbol{\tau}_i = {}^B\boldsymbol{\tau}_i$. 
    
    We assume that the hanging object always lies on a vertical plane $\mathcal{P}$ and does not overlap with itself. The rotational dynamics in \eqref{eq:euler} is not affected by the object, so an attitude controller that guarantees global stability, such as the geometric controller~\cite{5717652}, drives the vehicles to a stable trajectory tracking in the rotational dynamics.
    However, the unknown object's force $\boldsymbol{f}_i^o$ is not negligible and can lead to instability in the translational dynamics. 
    % In this paper, we focus on adapting to $\boldsymbol{f}_i^o$.
    
    \begin{problem*}[Trajectory tracking for bendable object transportation]
        % \label{pro:bendable}
        Given two vehicles connected to the two endpoints of an unknown bendable object with passive spherical joints, 
        design a control policy, $\boldsymbol{f}_i = \boldsymbol{u}_i$, to track a desired trajectory while adapting to the unknown force of the object, $\boldsymbol{f}_i^o$.
    \end{problem*}

    There are three major challenges in addressing the Problem. 
    First, the object's response to a vehicle's force is unknown and can cause instability.
    %Aggressive exploration-based methods may lead to uncertain elastic forces, causing the system to destabilize.
    \replytofive{Second, the lack of direct sensor measurements of the bendable force requires a force observer, of which the noise may harm the trajectory tracking performance.} 
    %An approximation method must be implemented in order to achieve controlled manipulation of the object. 
    Third, the low-latency requirement from the vehicles advocates a real-time method that is computationally light.

    \replytofour{The force compensation for bending the object is critical to trajectory tracking quality since the object force $\boldsymbol{f}^o_i$ directly influences the translational dynamics, as shown in~\eqref{eq:newton}. Thus, the primary objective is to design a control policy~$\boldsymbol{u}_i$ such that it adapts to the unknown bending force while tracking a trajectory. A key intuition about bendable objects is that a greater deformation requires a larger bending force, linking the force to the vehicles' states. Therefore, without an explicit model of the bendable object, the aerial vehicles can adapt by estimating the unknown force using their states, and use an acceleration observer to continuously update the estimation, and further refine the adaptation.}
    % Therefore, without an explicit model of the bendable object, the aerial vehicles can estimate their state using an acceleration observer and continuously refine the adaptation.}

\section{Adaptive Controller}
    We approach the Problem using adaptive control~\cite{aghajanzadeh2022adaptive}
    and model the object's force, $\boldsymbol{f}_i^o$, as a function of the vehicle states.

    % In order to solve Problem~\ref{pro:bendable}, we apply an approach based on adaptive control similar to~\cite{aghajanzadeh2022adaptive}
    % so that the vehicles drive the object endpoint to follow desired trajectories based on the approximation of $\boldsymbol{f}_i$.
    
    \subsection{\replytoall{Trajectory-tracking Control via Force Adaptation}}
    \label{sec:xy_feature}
    We estimate the force $\boldsymbol{f}_i^o$ using recursive least-square approximation~\cite{ioannou2012robust}.
    Since the bendable object force depends on its endpoint displacement $\boldsymbol{r}$, and the positional constraints ensure that the object curve remains in the same plane, we \replytoall{model an approximation function that maps} the displacement to the required force. 

    We \replytofive{assume} that the object's plane, $\mathcal{P}$, is \replytofive{vertical}, \textit{i.e.,} $\boldsymbol{z}_\mathcal{P} = \boldsymbol{z}\in\mathcal{P}$. \replytofive{This assumption allows us to approximate the orientation of $\mathcal{P}$ with only the knowledge of the endpoint positions of the rod, which is suitable for the scenario where the ``swinging'' behavior caused by the vehicle motion is neglectable.}
    The displacement vector~$\boldsymbol{r}$ determines the unit normal vector of $\mathcal{P}$, $\boldsymbol{y}_\mathcal{P} = \frac{\boldsymbol{r}\times\boldsymbol{z}_\mathcal{P}}{\Vert\boldsymbol{r}\times\boldsymbol{z}_\mathcal{P}\Vert}$. 
    The horizontal unit basis vector of $\mathcal{P}$ is $\boldsymbol{x}_\mathcal{P} = \boldsymbol{y}_\mathcal{P}\times\boldsymbol{z}_\mathcal{P}$. 
    Therefore, the orientation of $\mathcal{P}$ with respect to the world frame is $\boldsymbol{R}_\mathcal{P} = \begin{bmatrix}
        \boldsymbol{x}_\mathcal{P} & \boldsymbol{y}_\mathcal{P} & \boldsymbol{z}_\mathcal{P}
    \end{bmatrix}$. 
    %%%%%%%%%%%%%%%%%%%%%%%%%%%%%%%%%%%%%%%%%%%%
    \begin{figure}[t]
        \centering
        \includegraphics[width=\linewidth]{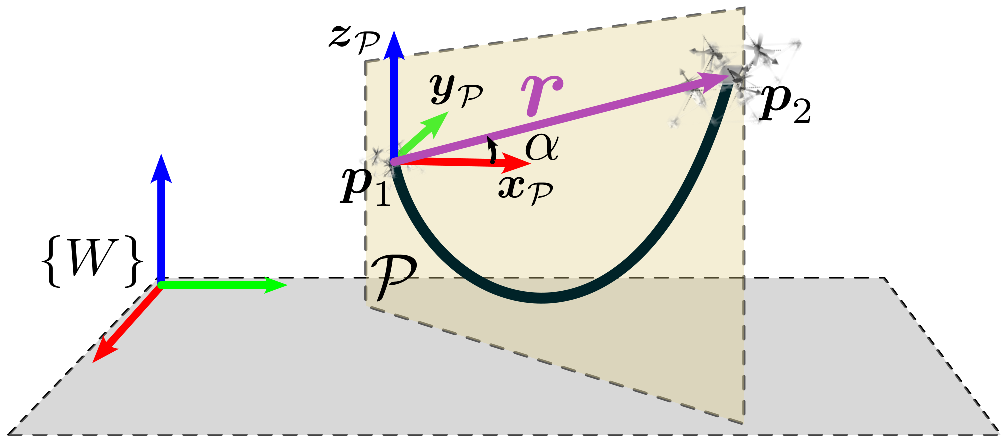}
        \caption{The strip lies in a vertical plane $\mathcal{P}$, the displacement of its two endpoints being $\boldsymbol{r}$.
        } 
        \label{fig:rod_xz}
    \end{figure}
    %%%%%%%%%%%%%%%%%%%%%%%%%%%%%%%%%%%%%%%%%%%%
    The displacement vector on~$\mathcal{P}$ is then $\boldsymbol{r}_\mathcal{P} = \boldsymbol{R}_\mathcal{P}^\top\boldsymbol{r}$. %, in which the second element is always $0$. \

    The object's force projected on the plane $\mathcal{P}$ is denoted by $\boldsymbol{f}^{o}_{i,\mathcal{P}}$.
    Then, we denote the estimator of $\boldsymbol{f}^o_{i,\mathcal{P}}$ by $\boldsymbol{\hat{f}}^{o}_{i,\mathcal{P}}$, of which the projection in the world frame is $\boldsymbol{\hat{f}}^{o}_{i} = \boldsymbol{R}_\mathcal{P}\boldsymbol{\hat{f}}^{o}_{i,\mathcal{P}}$. 
    % \fixme{we need to be consistent with the super indexes, if we use $^o$ for object's force, we should keep it on all the symbols}\
    We model the force $\boldsymbol{f}^{o}_{i,\mathcal{P}}$ as \replytofive{a mapping of a nonlinear} feature vector~$\boldsymbol{\phi}$ that depends on the vehicles' state, 
    \begin{equation}
        \boldsymbol{f}^{o}_{i,\mathcal{P}} = \boldsymbol{W}_i^\top\boldsymbol{\phi},
        \label{eq:approx_xz}
    \end{equation}
    where $\boldsymbol{W}_i\in\mathbb{R}^{\left(4n + 1\right)\times3}$ is the weight matrix. 
    \replytofive{
    Since the deformation and the bending force, is smooth and continuous on a closed interval, a polynomial of the endpoint displacement~$\boldsymbol{r}_\mathcal{P}$ uniformly approximates the function as accurately as desired, based on the Stone-Weierstrass Theorem~\cite{de1959stone}.    
    We use an $n$-th order polynomial feature of nonzero elements in $\boldsymbol{r}_\mathcal{P}$ and~$\boldsymbol{\dot r}_\mathcal{P}$,} 
    \begin{equation}
        \boldsymbol{\phi} = 
        \left[ r_x^n, r_x^{n-1}, \dots, r_x, 
        r_z^n, \dots, r_z, 
        {\dot r_x^n}, \dots, {\dot r_x}, 
        {\dot r_z^n}, \dots, {\dot r_z}, 
        1\right]^\top,
        \label{eq:feature_vector_xz}
    \end{equation}
    where $r_x$, $r_z$ are the first and third element of $\boldsymbol{r}_\mathcal{P}$. $\dot r_x$, $\dot r_z$ are the first and third element of $\boldsymbol{\dot r}_\mathcal{P}$, respectively. 
    \replytofive{We highlight that the use of polynomials requires fewer terms than other approximation functions such as neural networks. }
    %\replytofive{Since all components of $\boldsymbol{r}_\mathcal{P}$ for a non-extensible bendable object are bounded, and the response from the bendable object is continuous unless a critical point such as snapping is hit, by Stone-Weierstrass Theorem~\cite{de1959stone}, we design a polynomial feature vector for approximation.}
    We use continuous Recursive Least-Square algorithm~\cite{diniz1997adaptive,haykin2002adaptive} to update the approximation, $\boldsymbol{\hat{W}}_i$, of the unknown weight matrix $\boldsymbol{W}_i$, so that the estimator 
    \begin{equation}
        \boldsymbol{\hat f}^{o}_{i,\mathcal{P}} = \boldsymbol{\hat W}_i^\top\boldsymbol{\phi}
        \label{eq:estimated_force}
    \end{equation} 
    converges to $\boldsymbol{f}^{o}_{i,\mathcal{P}}$ as the sample size becomes infinite.
    Since the object's force cannot be easily measured directly, 
    we use \replytoall{an acceleration-based force observer to obtain it based on the vehicle dynamics in \eqref{eq:newton}},
    % ,
    \begin{equation}
        \boldsymbol{f}_i^o = m\ddot{\boldsymbol{p}_i} + mg\boldsymbol{z} - \boldsymbol{u}_{i},\label{eq:inversenewton}
    \end{equation}
    \noindent
    where the vehicle's force $\boldsymbol{u}_i$ can be computed based on the actuator inputs, and the vector~$\ddot{\boldsymbol{p}_i}$ can be measured. \replytoall{In practice, the measurement is noisy due to the nature of taking double derivatives on the measured position and noisy measurements from the inertial sensor. 
    In addition, the measurement is only available after motor actuation and the states are observed. 
    To ensure the trajectory tracking controller addresses the bendable force while taking action, we estimate the force based on the expected states before measurement, and update the estimation based on the measurement.}
    
    We initialize the weight matrix as a zero matrix, $\boldsymbol{\hat W}_i(0) = \boldsymbol{0}$, and the covariance as the identity matrix, $\boldsymbol{P}_i(0) = \boldsymbol{I}$.
    The update rule given by the continuous-time recursive least-square algorithm with a forgetting factor~\cite{ioannou2012robust} is,
    \begin{eqnarray}
        \boldsymbol{\dot{\hat {W}}}_i &=& \boldsymbol{P}_i\boldsymbol{\phi}\, \boldsymbol{\epsilon}^\top_i\label{eq:update_weight}\\
        \boldsymbol{\dot P}_i &=& \lambda\boldsymbol{P}_i - \boldsymbol{P}_i\boldsymbol{\phi}_i\boldsymbol{\phi}_i^\top\boldsymbol{P}_i\label{eq:update_corm}
    \end{eqnarray}
    where the force estimation error in $\mathcal{P}$ is
    \begin{equation}
    \boldsymbol{\epsilon}_i = \boldsymbol{R}_\mathcal{P}^\top\boldsymbol{f}_i^o - \boldsymbol{\hat{f}}_{i,\mathcal{P}}^o.
        \label{eq:epsilon_error}
    \end{equation}
    % \boldsymbol{\epsilon}_i = \boldsymbol{R}_\mathcal{P}^\top\boldsymbol{f}_i^o - \boldsymbol{\hat W}^\top_{i}\boldsymbol{\phi}.
    % \begin{equation}
    %     \boldsymbol{L}_{i}(t) = \boldsymbol{P}_i(t-\delta t)\boldsymbol{\phi}\left(\lambda + \boldsymbol{\phi}^\top\boldsymbol{P}_{i}(t-\delta t)\boldsymbol{\phi}\right)^{-1}.
    %     \label{eq:gain_vector}
    % \end{equation}
    % \noindent
    % Then we update the weight matrix,
    % \begin{equation}
    %     \boldsymbol{\hat W}_{i}(t) = \boldsymbol{\hat W}_{i}(t-\delta t) + \boldsymbol{L}_{i}(t)\boldsymbol{\epsilon}_i^\top,
    %     \label{eq:update_weight}
    % \end{equation}
    % based on  The recursively updated inverse correlation matrix 
    % \begin{equation}
    %     \boldsymbol{P}_{i}(t) = \frac{1}{\lambda}\left(\left(\boldsymbol{I} -\boldsymbol{L}_{i}(t)\boldsymbol{\phi}^\top\right)\boldsymbol{P}_{i}(t-\delta t)\right)
    %     \label{eq:update_corm}
    % \end{equation}
    % \noindent
    % is initialized as a scalar multiplication with the identity matrix $\boldsymbol{P}_{i, 0} = \delta\boldsymbol{I}$ for a large $\delta\gg1$, where $\boldsymbol{I}$ is the identity matrix, and $0 < \lambda < 1$ is the forgetting factor for the recursive updates. 
    Using the updated weight matrix, the vehicles track the desired trajectories while adapting to the elastic force simultaneously.
    
    % Since it is able to track an arbitrary attitude using the geometric controller~\cite{5717652,5980409}, and the object disturbance torque $\boldsymbol{\tau}^e_i = \boldsymbol{0}$, 
    % the attitude control is stable. Therefore,
    % We consider the vehicle force generation in the world frame,
    The objective of our control policy is to track each endpoint of the object to a desired trajectory $\boldsymbol{p}^d_i$ by driving the position error, $\boldsymbol{e}_i = \boldsymbol{p}_i^d - \boldsymbol{p}_i$, and velocity error, $\boldsymbol{\dot e}_i = \boldsymbol{\dot p}_i^d - \boldsymbol{\dot p}_i$, to zero, while adapting to the object's unknown force. Our proposed control policy \replytoseven{ combines the proportional-derivative trajectory tracking controller and the force adaptation,}
    \begin{equation}
        \boldsymbol{u}_{i} = k_p\boldsymbol{e}_i + k_d\boldsymbol{\dot e}_i + mg\boldsymbol{z} + m\boldsymbol{\ddot p}^d - \boldsymbol{\hat f}^o_i,
        \label{eq:control_input}
    \end{equation}
    where $k_p, k_d > 0$ are scalar proportional and derivative gains for the errors. 
    The trajectory desired acceleration $\boldsymbol{\ddot p}^d$ is the feedforward term. We estimate the object's force $\boldsymbol{\hat f}^o_{i,\mathcal{P}}$ on the~$\mathcal{P}$ plane using the least-square approximation and project it back to the world frame to obtain $\boldsymbol{\hat f}^o_{i}$. 
    \replytoall{
    Although a PID controller can use the accumulated error to compensate for the bending force, it would work when the bending force is constant, otherwise, the changes in the bending force can lead to oscillations, delays, and instability.    
    % In comparison to a PID controller which compensates for the accumulative errors caused by the unknown elastic force through an integral term on top of the PD-based trajectory tracking, our proposed force adaptation, in replacement of the integral term, is invariant to the $\mathcal{P}$ rotation, and takes consideration of the physical intuition of elastic objects, which associates the required bending force with the endpoint displacement.
    }

    \replytoseven{For the approximated weight matrix $\boldsymbol{\hat W}_i$ to converge to the actual matrix $\boldsymbol{W}_i$, the input $\sum_{q=t}^{t+N-1}\boldsymbol{\phi}(q)\boldsymbol{\phi}(q)^\top$ must be persistently excited (PE)~\cite{ioannou2012robust}. To satisfy PE, we can design trajectories with diverse motion, typically by using periodic or aperiodic signals that span a broad frequency range. Alternatively, we can superimpose oscillatory components at varying frequencies onto the original trajectories. Consequently, the collected data becomes informative enough to maintain a full-rank sum of outer products, guaranteeing the controller's convergence and stability.}
    % Based on the theory estimation  \cite{}
    % For our system to satisfy the persistence of excitation condition, \textit{i.e.,} to make $\sum_{q=t}^{t+N-1}\boldsymbol{\phi}(q)\boldsymbol{\phi}(q)^\top$ full-rank and bounded, we ensures the desired trajectories are superpositioned with oscillations in position at different frequencies.

    \subsection{Stability}
    \label{sec:stability_proof}
    \replytofour{Based on the assumption that our force estimation model in~\eqref{eq:approx_xz} is sufficiently accurate to an arbitrary precision of choice according to Stone-Weierstrass Theorem,} we study the stability of the system described in~\eqref{eq:newton} using our control input in~\eqref{eq:control_input} in the task of trajectory tracking, 
    % Given the vehicle under disturbance of the object force, the positional errors, and the adaptive force compensation by considering~\eqref{eq:control_input} into~\eqref{eq:newton}, the linear dynamics becomes 
    \begin{equation}
        m\ddot{\boldsymbol{p}_i} = k_p\boldsymbol{e}_i + k_d\boldsymbol{\dot e}_i + m\boldsymbol{\ddot p}^d + \boldsymbol{f}_i^o - \boldsymbol{\hat f}^o_i,\label{eq:new_newton}
    \end{equation}
    % With the assumption that \replytofive{an approximation function that maps} the feature vector can accurately describe the object's force in the form 
    where $\boldsymbol{f}_i^o = \boldsymbol{R}_\mathcal{P}\boldsymbol{W}_i^\top\boldsymbol{\phi}$. We rearrange the terms and include the force estimation in $\mathcal{P}$ with~\eqref{eq:approx_xz} and~\eqref{eq:estimated_force} to obtain % desired acceleration and the error of weight matrix estimation in~\eqref{eq:new_newton}, we obtain the error dynamics,    
    \begin{equation}
        m\boldsymbol{\ddot e}_i = - k_d\boldsymbol{\dot e}_i - k_p\boldsymbol{e}_i + \boldsymbol{R}_\mathcal{P}\boldsymbol{\Tilde{W}}_i^\top\boldsymbol{\phi}
        \label{eq:error_dynamics},
    \end{equation}
    where $\boldsymbol{\Tilde{W}}_i = \boldsymbol{W}_i - \boldsymbol{\hat{W}}_i$ is the weight matrix error, which we can use to rewrite the force estimation error as
    \begin{equation}
        \boldsymbol{\epsilon}_i = \boldsymbol{\Tilde{W}}_i^\top\boldsymbol{\phi}.
        \label{eq:f_error_estimation}
    \end{equation}
    We define a candidate Lyapunov function considering the error dynamics and the estimation error,
    \begin{eqnarray}
        V_i &=& V_i^p + V_i^W, \quad\text{where}\\\label{eq:lyapunov}
        V_i^p &=& \frac{1}{2}m\boldsymbol{\dot e}_i^\top\boldsymbol{\dot e}_i + 
        \frac{1}{2}\boldsymbol{e}_i^\top k_p\boldsymbol{e}_i,\label{eq:v1}\\
        % \fixme{\text{why is an scalar in the middle of vectors?}} \\ because we want to eliminate some terms in the derivative
        V_i^W &=& \text{tr}(\boldsymbol{\Tilde{W}}_i^\top\boldsymbol{P}_i^{-1}\boldsymbol{\Tilde{W}}_i).\label{eq:v2}
    \end{eqnarray}
    The operator $\text{tr}(\cdot)$ denotes the trace of a matrix. Note that~$V_i^p$ is associated with the tracking error and $V_i^W$ is with the estimation error. Since the proportional gain $k_p$ and the inverse covariance matrix $\boldsymbol{P}$ are both positive definite, the Lyapunov candidate is positive definite. If and only if the position error $\boldsymbol{e}_i = \boldsymbol{0}$, the position error derivative $\boldsymbol{\dot e}_i = \boldsymbol{0}$, and the weight matrix error $\boldsymbol{\Tilde{W}}_i = \boldsymbol{0}$ are zero, the candidate $V_i = 0$ is zero. 
    
    The derivative of the Lyapunov candidate,
    $\dot V_i = \dot V_i^p + \dot V_i^W$ contains two major components. 
    First, we simplify the derivative of the estimation component,
    \begin{equation}
        \dot V_i^W = 2\text{tr}(\boldsymbol{\Tilde{W}}_i^\top\boldsymbol{P}_i^{-1}\boldsymbol{\dot{\Tilde{W}}}_i) + \text{tr}(\boldsymbol{\Tilde{W}}_i^\top\frac{d}{dt}(\boldsymbol{P}_i^{-1})\boldsymbol{\Tilde{W}}_i).\label{eq:v2dot}
    \end{equation}
    The weight matrix update rule in~\eqref{eq:update_weight} induces $\boldsymbol{\dot{\Tilde{W}}}_i = -\boldsymbol{\dot{\hat{W}}}_i = -\boldsymbol{P}_i\boldsymbol{\phi}\boldsymbol{\epsilon}^\top_i$. Assuming $\lambda = 0$ for the recursive least-squares without forgetting factor, the covariance matrix update rule in~\eqref{eq:update_corm} gives $\boldsymbol{\dot P}_i = -\boldsymbol{P}_i\boldsymbol{\phi}_i\boldsymbol{\phi}_i^\top\boldsymbol{P}_i$. Combined with the identity $\frac{d}{dt}(\boldsymbol{P}_i^{-1}\boldsymbol{P}_i)\equiv\boldsymbol{0}$, which gives $\frac{d}{dt}(\boldsymbol{P}_i^{-1}) = -\boldsymbol{P}_i^{-1}\boldsymbol{\dot P}_i\boldsymbol{P}^{-1}_i$, we simplify~\eqref{eq:v2dot} into
    \begin{eqnarray}
        \dot V_i^W &=& -2\text{tr}(\boldsymbol{\Tilde{W}}_i^\top\boldsymbol{\phi}\boldsymbol{\epsilon}_i^\top) - \text{tr}(\boldsymbol{\Tilde{W}}_i^\top\boldsymbol{P}_i^{-1}\boldsymbol{\dot P}_i\boldsymbol{P}_i^{-1}\boldsymbol{\Tilde{W}}_i)\nonumber\\
        &=& -2\text{tr}(\boldsymbol{\Tilde{W}}_i^\top\boldsymbol{\phi}\boldsymbol{\epsilon}_i^\top)\nonumber\\
        && -\text{tr}(\boldsymbol{\Tilde{W}}_i^\top\boldsymbol{P}_i^{-1}(-\boldsymbol{P}_i\boldsymbol{\phi}_i\boldsymbol{\phi}_i^\top\boldsymbol{P}_i)\boldsymbol{P}_i^{-1}\boldsymbol{\Tilde{W}}_i)\nonumber\\
        &=& -2\text{tr}(\boldsymbol{\epsilon}_i\boldsymbol{\epsilon}_i^\top) + \text{tr}(\boldsymbol{\epsilon}_i\boldsymbol{\epsilon}_i^\top)\nonumber\\
        &=& - \boldsymbol{\epsilon}_i^\top\boldsymbol{\epsilon}_i < 0,\label{eq:v2dotnew}
    \end{eqnarray}
    which is negative definite. If and only if the force estimation error $\Vert\boldsymbol{\epsilon}_i\Vert = 0$ is zero, this derivative component of the Lyapunov function is zero. Combined with the positive definiteness of $V_i^W$ from~\eqref{eq:v2}, it shows the global asymptotic attraction of the estimation error to zero.
    The second component is the derivative related to tracking errors,
    \begin{equation}
        \dot V_i^p = m\boldsymbol{\dot e}_i^\top\boldsymbol{\ddot e}_i +\boldsymbol{e}_i^\top\ k_p\boldsymbol{\dot e}_i.\label{eq:v1dot}
    \end{equation}

    \noindent
    Substituting $\boldsymbol{\ddot e}$ with the error dynamics in~\eqref{eq:error_dynamics}, we obtain 
    \begin{eqnarray}
        \dot V_i^p &=& \boldsymbol{\dot e}_i^\top\left(\boldsymbol{R}_\mathcal{P}\boldsymbol{\Tilde{W}}_i^\top\boldsymbol{\phi} - k_p\boldsymbol{e}_i - k_d\boldsymbol{\dot e}_i\right) + \boldsymbol{e}_i^\top k_p\boldsymbol{\dot e}_i\nonumber\\
        &=& \boldsymbol{\dot e}_i^\top\boldsymbol{R}_\mathcal{P}\boldsymbol{\Tilde{W}}_i^\top\boldsymbol{\phi} - \boldsymbol{\dot e}_i^\top k_d\boldsymbol{\dot e}_i\nonumber\\
        &=& \boldsymbol{\dot e}_i^\top\boldsymbol{R}_\mathcal{P}\boldsymbol{\epsilon}_i - \boldsymbol{\dot e}_i^\top k_d\boldsymbol{\dot e}_i.
        % &\leq& \Vert\boldsymbol{\dot e}_i\Vert\Vert\boldsymbol{\epsilon}_i\Vert - k_d\Vert\boldsymbol{\dot e}_i\Vert^2
        \label{eq:v1dotnew}
    \end{eqnarray}
    The Lyapunov function as a result of the derivation of its two components in \eqref{eq:v2dotnew} and \eqref{eq:v1dotnew} is,
    \begin{eqnarray}
        \dot V_i &=& \dot V_i^W + \dot V_i^p\nonumber\\
        &=& -\boldsymbol{\epsilon}_i^\top\boldsymbol{\epsilon}_i + \boldsymbol{\dot e}_i^\top\boldsymbol{R}_\mathcal{P}\boldsymbol{\epsilon}_i - \boldsymbol{\dot e}_i^\top k_d\boldsymbol{\dot e}_i.\label{eq:dotlyapunovcandidate}
    \end{eqnarray}
    The right-hand side of~\eqref{eq:dotlyapunovcandidate} is composed of two negative squares and $\boldsymbol{\dot e}_i^\top\boldsymbol{R}_\mathcal{P}\boldsymbol{\epsilon}_i$. Since the rotation matrix $\boldsymbol{R}_\mathcal{P}$ changes only the direction of the vector $\boldsymbol{\epsilon}_i$, the maximum value of $\boldsymbol{\dot e}_i^\top\boldsymbol{R}_\mathcal{P}\boldsymbol{\epsilon}_i$ is bounded by $\Vert\boldsymbol{\dot e}_i\Vert\Vert\boldsymbol{\epsilon}_i\Vert$ when $\boldsymbol{\dot e}_i$ is co-linear with $\boldsymbol{R}_\mathcal{P}\boldsymbol{\epsilon}_i$, \textit{i.e.,}
    \begin{eqnarray}
        \dot V_i &\leq& -\Vert\boldsymbol{\epsilon}_i\Vert^2 + \Vert\boldsymbol{\dot e}_i\Vert\Vert\boldsymbol{\epsilon}_i\Vert - k_d\Vert\boldsymbol{\dot e}_i\Vert^2\label{eq:dotlyapunovcandidatepart1}
    \end{eqnarray}
    We can reorganize the right-hand side of~\eqref{eq:dotlyapunovcandidatepart1} into the summation of two negative square terms,
    \begin{equation}
        \dot V_i \leq - (\Vert\boldsymbol{\epsilon}_i\Vert - \frac{1}{2}\Vert\boldsymbol{\dot e}_i\Vert)^2 - (k_d - \frac{1}{4})\Vert\boldsymbol{\dot e}_i\Vert^2 < 0,
        % &\leq& - (k_d - \frac{1}{4})\Vert\boldsymbol{\dot e}_i\Vert^2
        \label{eq:dotlyapunovcandidatepart2}
    \end{equation}
    which is negative definite for $k_d > \frac{1}{4}$. If and only if both the velocity tracking error $\boldsymbol{\dot e}_i = \boldsymbol{0}$ and the force estimation error $\boldsymbol{\epsilon}_i = \boldsymbol{0}$ are zero, the Lyapunov derivative is zero. 
    As a result, we have the Lyapunov function $V_i$ positive definite, and its derivative $\dot V_i$ negative definite for $k_d > \frac{1}{4}$. When the tracking errors and the force estimation error are both zero, $V_i = 0$. Since the feature vector $\boldsymbol{\phi}$ is a polynomial of the bounded endpoint displacement, the magnitude of the weight matrix error is bounded. Likewise, the magnitude of the position and velocity errors for the task of trajectory-tracking is bounded. Thus, the system of aerial vehicles and the bendable object is \emph{asymptotically stable} in tracking the endpoint trajectories while compensating for the object's force.% with the proposed adaptive controller.

    \replytoall{It is noted that depending on the order of polynomial and the model mismatch unaccounted for such as the misalignment from the vehicle CoM and the connection point to the bendable object, the force estimation model is not accurate in practice. Consequently, the force estimation error $\boldsymbol{\boldsymbol{\epsilon}_i}^\top\boldsymbol{\boldsymbol{\epsilon}_i}$ in $\mathcal{P}$ will stabilize around $0$, resulting in a residual trajectory tracking error. Though a feature vector of higher polynomial order offers lower estimation errors, it also increases the computation overhead as the size of $\boldsymbol{\hat W}_i$ increases. An alternative to lower the estimation error is to adopt a different feature vector which incorporates the physical insights of a bendable objects.}

    \subsection{Physical Insights of a Bendable Object}
    \begin{figure}[t]
        \centering
        \includegraphics[width=\linewidth]{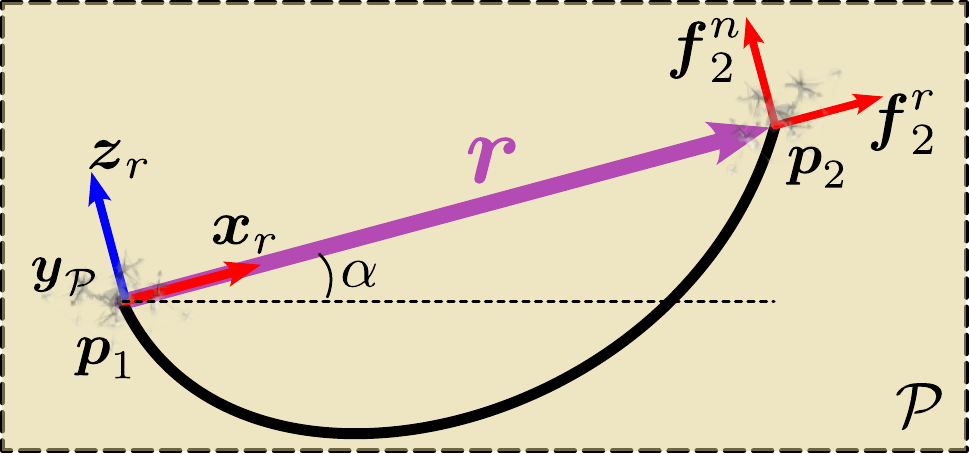}
        \caption{The object force in parallel and perpendicular to displacement vector in the plane $\mathcal{P}$. The force vectors on vehicle 1 omitted for illustration clarity.}
        \label{fig:rod_nr}
    \end{figure}
    The most fundamental physical characteristic of a bendable object is its tendency to recover its endpoint displacement $r$ to the natural value $r_0$. When the object is bent, $r \neq r_0$ leads the object to resist bending, resulting in the elastic force received by the vehicles. During transportation, the vehicles also need to take into account the force due to gravity, which is nontrivial because of its unknown mass distribution. The gravity compensation of each vehicle depends on the leaning of the object, $\alpha$, which is the angle between $\boldsymbol{r}$ and the horizontal plane,
    \begin{equation}
        \alpha = \arctan \left( {\frac{\boldsymbol{r}\cdot\boldsymbol{z}_\mathcal{P}}{\boldsymbol{r}\cdot\boldsymbol{x}_\mathcal{P}}} \right). \label{eq:alpha}\\
    \end{equation}
    These insights motivate us to construct a new feature vector for improved force estimation. 
    We first decompose the total force received by the vehicles from the object in parallel and perpendicular to the displacement vector $\boldsymbol{r}$, that is, $\boldsymbol{f}_{i}^{r}$ and $\boldsymbol{f}_{i}^{n}$ at endpoint $i$, as shown in Fig.~\ref{fig:rod_nr}. As such, the external force applied by the object is $\boldsymbol{f}^o_i = \boldsymbol{f}_i^r + \boldsymbol{f}_i^n$. 
    To ensure that this modification is consistent with the methods introduced in Section.~\ref{sec:xy_feature}, we find the unit vector of the displacement $\boldsymbol{x}_r = \frac{\boldsymbol{r}}{\Vert\boldsymbol{r}\Vert}$, and the normal vector of $\boldsymbol{x}_r$ in $\mathcal{P}$, $\boldsymbol{z}_r = \text{Rot}(\boldsymbol{y}_\mathcal{P}, -\frac{\pi}{2})\boldsymbol{x}_r$ by applying Rodrigues' rotation formula.
    % , we can express the external force $\boldsymbol{f}^o_i = \lambda_i^r \boldsymbol{x}_r + \lambda_i^n \boldsymbol{z}_r$. 
    Thus, we obtain the rotation matrix $\boldsymbol{R}_o\triangleq\begin{bmatrix}\boldsymbol{x}_r & \boldsymbol{y}_{\mathcal{P}} & \boldsymbol{z}_r\end{bmatrix}$ using the three orthogonal unit vectors. This rotation matrix, combined with a selected origin at $\boldsymbol{p}_1$, defines a new frame $\{O\}$, of which the $xz$-plane is identical to $\mathcal{P}$ with a different set of basis vectors. By projecting $\boldsymbol{f}^o_i$ into $\{O\}$, we obtain $\boldsymbol{f}^o_{i,o} = \boldsymbol{R}_o^\top\boldsymbol{f}^o_{i}$, of which the first and third elements are the magnitudes of $\boldsymbol{f}_{i}^{r}$ and $\boldsymbol{f}_{i}^{n}$. We define the feature vector
    \begin{equation}
    \footnotesize
        \boldsymbol{\phi}^b =
        \left[ r^n, r^{n-1}, \dots, r, 
        {\dot r^n}, \dots, {\dot r}, 
        \cos{n\alpha}, \dots, \cos{\alpha},
        \sin{n\alpha}, \dots, \sin{\alpha},
        1\right]^\top.
    \end{equation}
    \noindent
    We model the object force in $\{O\}$ as a linear function of the new feature vector $\boldsymbol{\phi}^b$, \textit{i.e.,} $\boldsymbol{f}^o_{i,o} = \boldsymbol{W}^b_i\boldsymbol{\phi}^b$. The estimator of~$\boldsymbol{f}^o_{i,o}$ is denoted by $\boldsymbol{\hat f}^o_{i,o} = \boldsymbol{\hat W}^b_i\boldsymbol{\phi}^b$ and $\boldsymbol{\hat W}^b_i\in\mathbb{R}^{(4n+ 1)\times3}$ is the associated estimation weight matrix.
    While the update rules remain the same as in~\eqref{eq:update_weight}, \eqref{eq:update_corm}, the estimation error is
    \begin{equation}
        \boldsymbol{\epsilon}_i^b = \boldsymbol{R}^\top_o\boldsymbol{f}^o_i - \boldsymbol{\hat f}^o_{i,o}.
        \label{eq:epsilon_ralpha}
    \end{equation}
    This new feature vector takes into consideration the physical characteristics of a bendable object without explicit knowledge of the model, thus generalizes over different bendable objects. The stability proof follows the same procedure as in Section.~\ref{sec:stability_proof} by replacing $\boldsymbol{R}_\mathcal{P}$ with $\boldsymbol{R}_o$.
    The performance of both feature vectors is evaluated in the next section.

\section{Experiments}
% \begin{figure}[t]
%     \centering
%     \includegraphics[width=\linewidth]{figures/exp1.png}
%     \caption{The quadrotors' trajectory for experiment 1.}
%     \label{fig:exp1}
% \end{figure}
We connect a quadrotor to each endpoint of a thin and long carbon-fiber strip and use the two quadrotors to drive the bendable object's endpoints to follow a desired trajectory to validate our methods. We measure the performance in terms of position tracking errors and compare the performance of our adaptive control to that of the standard PID control. We evaluate the methods using three different trajectories. For simplicity, we refer to the system of the two quadrotors and the bendable strip as ``the system'' in this section.

We use the quadrotor modules from the H-ModQuad project~\cite{9561016} as the aerial vehicle. Each module weighs $135$ g and has a payload capacity of over $100$ g. A ground station calculates and sends the trajectory waypoints to the quadrotors using the Crazyswarm~\cite{7989376} framework at $100$ Hz. The ground station executes the RLS calculation and sends the compensation force for bending to the quadrotors \replytoseven{consistently throughout an experiment}. The strip for the experiments is constructed as a composite of multiple carbon fiber strips of different weights, thicknesses, base shapes, and lengths to achieve asymmetry and nonhomogeneity.\footnote{The complete set of video and data recordings is available at \url{https://tinyurl.com/bendables}. %A summary video of the experiments is available at \url{https://www.youtube.com/watch?v=lhTVIWSjhvQ}.
}

\begin{figure}[t]
    \centering
    \includegraphics[trim={1cm 0cm 2cm 1cm},clip,width=\linewidth]{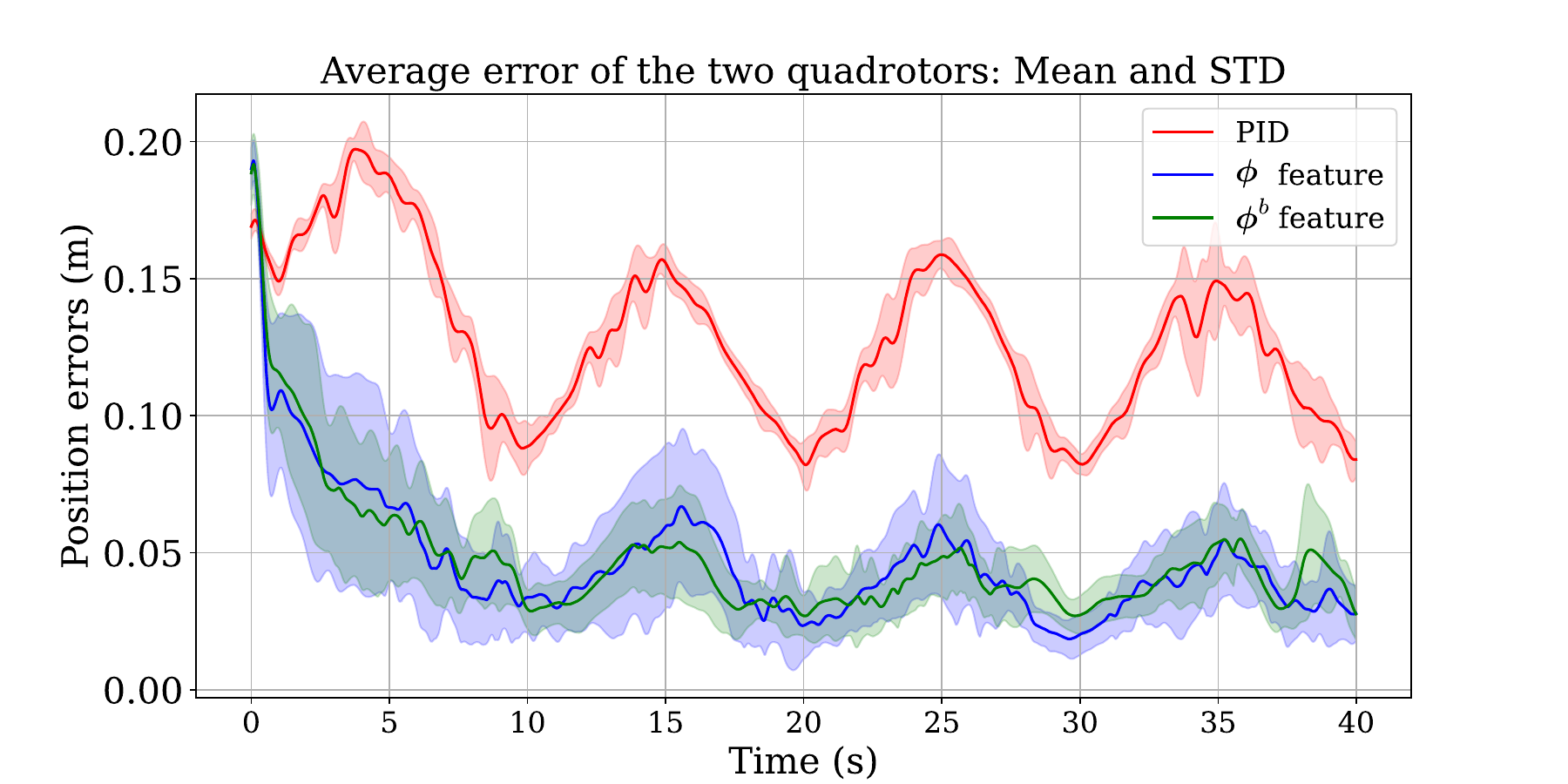}
    \caption{The mean and standard deviation of the average endpoint position errors v.s. time for tracking a varying endpoint distance.}
    \label{fig:convergence_time}
\end{figure}
\begin{figure}[t]
    \centering
    \includegraphics[trim={1cm 0cm 2cm 1cm},clip,width=\linewidth]{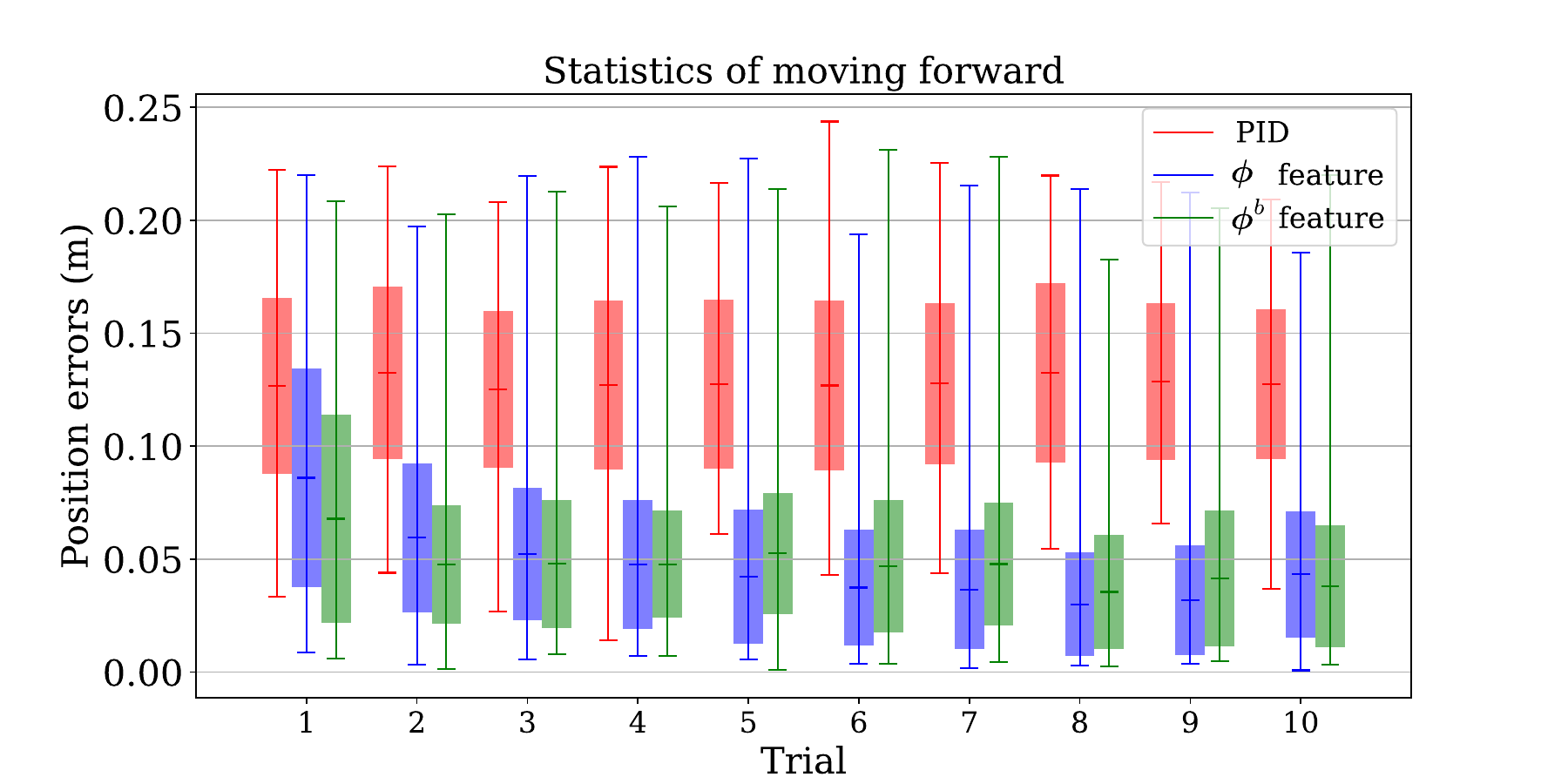}
    \caption{The mean and standard deviation of the average endpoint position errors v.s. trial for tracking a varying endpoint distance.}
    \label{fig:convergence_stats}
\end{figure}

\subsection{Trajectory-tracking Evaluation}{
    \label{sec:exp1}
    In the first experiment, we command the system to translate in $x$-direction at a constant speed while maintaining height and the $\mathcal{P}$ plane orientation, The endpoint distance of the strip varies sinusoidally, and the desired height of both endpoints oscillates at different frequencies with a magnitude of $0.05$ m.
    The strip has a curve length of $L_0 = 1.2$ m and a natural endpoint displacement $r_0 = 1.05$~m. During transportation, we command the quadrotors to oscillate the endpoint displacement between $0.8$ m and $0.4$ m. This trajectory evaluates the strength and convergence speed of the force compensation methods during strip bending. Using our adaptive controller, the system can quickly and consistently reduce tracking errors that are significantly smaller than the PID baseline. We run the trajectory $10$ times with each method for $40$ seconds each time. For adaptive controllers, we continue to update the RLS matrices, \textit{i.e.}, $\boldsymbol{W}_i$ and $\boldsymbol{P}_i$, across the trials. The mean and standard deviation (STD) of the average tracking errors of the two endpoints over time are shown in Fig.~\ref{fig:convergence_time}. 
    The mean and STD of the average tracking errors of the two endpoints over the $10$ trials are shown in~Fig.~\ref{fig:convergence_stats}.

    We highlight three observations. First, while the PID baseline consistently achieves a mean error around $0.13$ m with an STD of $0.03$ m, our adaptive controllers with both feature vectors achieve a lower mean error less than $0.1$ m with an STD of $0.04$ m during the first trial, and eventually achieve a \emph{significantly} lower mean error less than $0.05$ m with an STD of $0.02$ m. Second, since the PID controller compensates for the bendable force by accumulating the integral term, it does not comprehend the characteristics of the bendable strip, resulting in a high deviation in error. In contrast, our controller learns the bendable force by updating the weight matrices, thus achieving a lower deviation. Third, the \replytoseven{$\boldsymbol{\phi}^b$} feature vector provides a deeper understanding of the bendable characteristics, which leads to faster convergence and lower errors compared to using the \replytoseven{$\boldsymbol{\phi}$} feature vector. 
    \replytoall{In addition, we assume that a quadrotor connects to the object's endpoint at its CoM, and thus the bendable object applies zero torque on the quadrotors. However, in practice, slight deviations in the connection point from the quadrotor CoM create unexpected torques. These torques amplify errors both in force compensation magnitude and direction in the PID baseline, which results in higher tracking errors as bending the object more requires higher force.}
}

\begin{figure}[t]
        \centering
        \includegraphics[trim={1cm 0cm 2cm 1cm},clip,width=\linewidth]{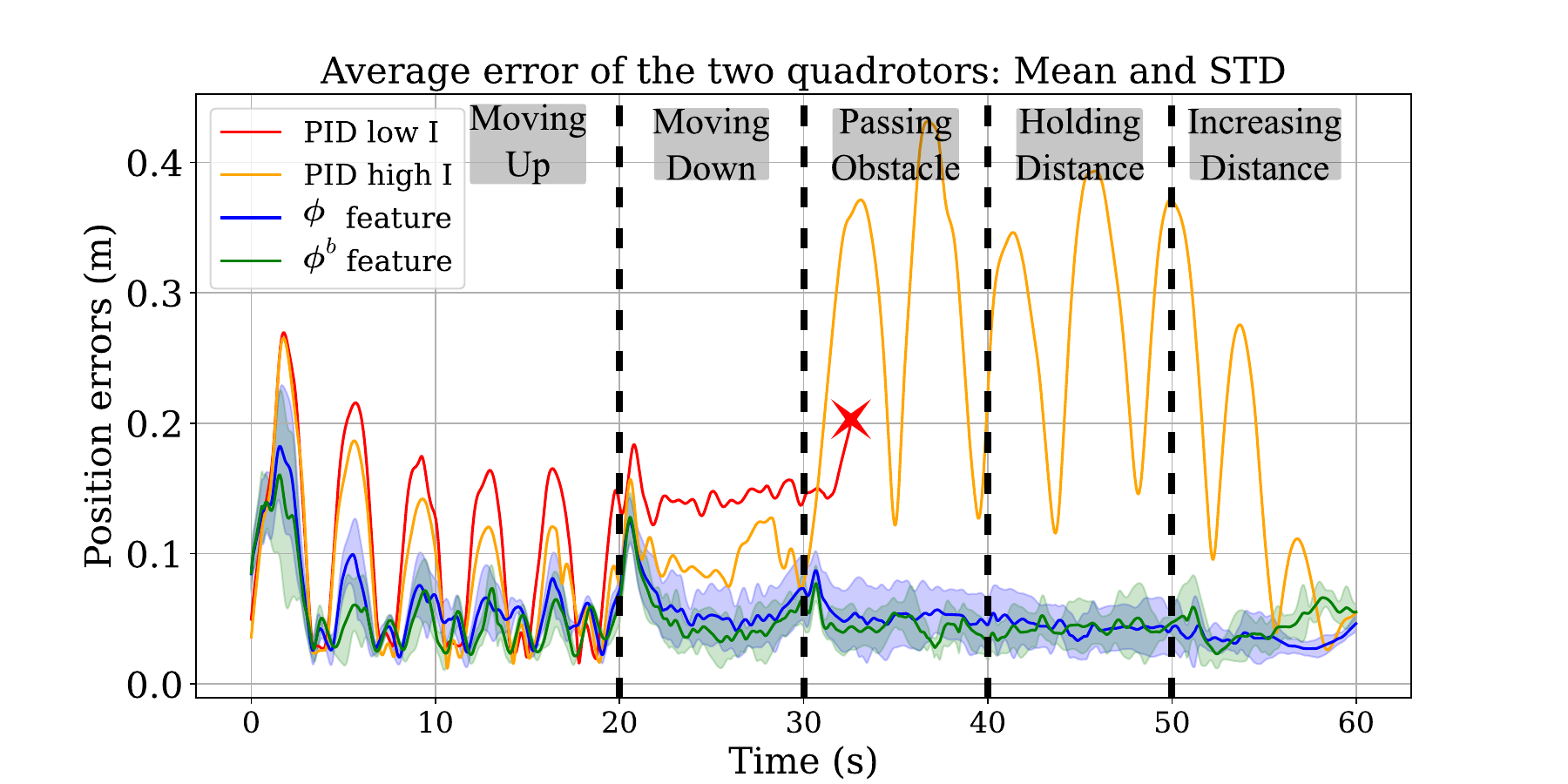}
        \caption{The mean and STD of the average endpoint position errors v.s. time for climbing up and passing one obstacle. The red cross marks where the PID controller with a low integral term fails to drive the strip through the obstacle.}
        \label{fig:one_obs_time}
    \end{figure}
    \begin{figure}[t]
        \centering
        \includegraphics[trim={1cm 0cm 2cm 1cm},clip,width=\linewidth]{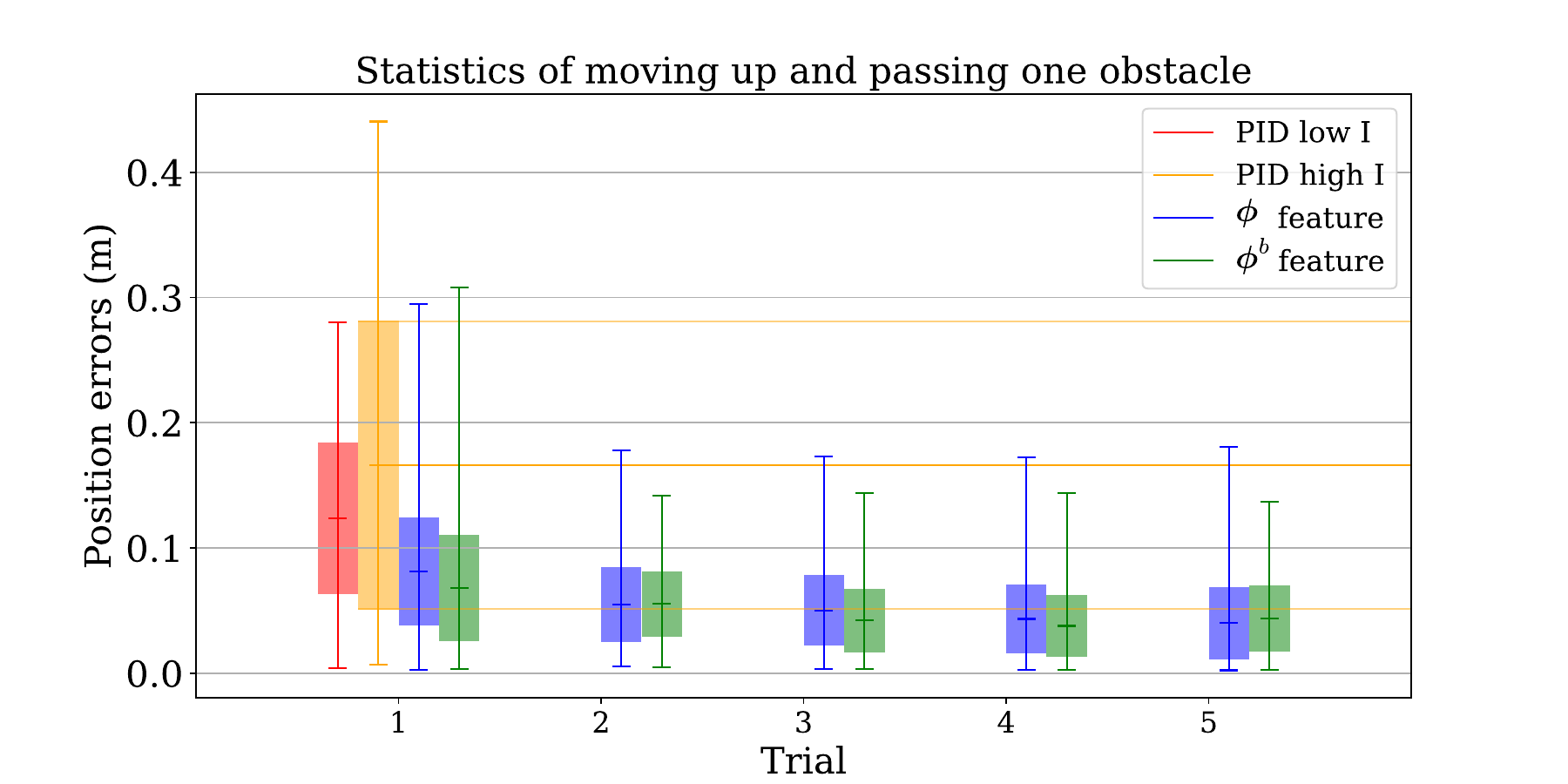}
        \caption{The mean and standard deviation of the average endpoint position errors v.s. trial for climbing up and passing one obstacle.}
        \label{fig:one_obs_trial}
    \end{figure}
\subsection{Simple Task: Passing Through a Window}{
    % \begin{figure}[t]
    %     \centering
    %     \includegraphics[width=\linewidth]{figures/exp2.png}
    %     \caption{The quadrotors' trajectory for experiment 2.}
    %     \label{fig:exp2}
    % \end{figure}
    In the second experiment, we design a trajectory to move forward-up, and then pass through a square window. It is composed of five parts. First, the system translates in $x$-direction at a constant speed, moves up, and maintains the $\mathcal{P}$-plane orientation at the same time until the system reaches a height of $1.8$ m. The upward movement is combined with a varying endpoint distance between $0.6$ m and $1.1$ m sinusoidally, and both endpoints change their relative heights to the system height sinusoidally at different frequencies with a magnitude of $0.05$ m. Second, the system maintains the endpoint displacement distance at $0.6$ m and moves down in the $z$-direction while maintaining the $\mathcal{P}$ orientation and the $x$-axis speed until the height reaches $0.7$ m. Third, the two quadrotors maintain the endpoint displacement while moving forward to pass through the square obstacle. Fourth, after going through the obstacle, the system stays static for $10$ s. Lastly, the system increases the displacement distance to the same value during takeoff, then lands, finishing the task.

    The results in terms of position errors are shown in Fig.~\ref{fig:one_obs_time} and~\ref{fig:one_obs_trial}. We run the PID baselines only once since their performance does not improve with repetitions. The PID baseline with the \replytoall{default low integral term for unconstrained quadrotors} fails to pass through the obstacle as error integration does not generate enough force to bend the strip timely, marked by a red cross in Fig.~\ref{fig:one_obs_time}. PID with \replytoall{twice as high of the default integral} term succeeds in passing the obstacle but struggles to damp the position error when the quadrotors are commanded to stay static while bending the strip. The adaptive controllers with both feature vectors complete the task and maintain the average position error below $0.05$ m after $3$ trials, with an STD less than $0.03$ m. The task lasts $60$~s, which is $20$ s longer than Exp.~\ref{sec:exp1}. The longer runtime allows the adaptive controllers to converge further to a lower error within the first trial. Similar observation as in~\ref{sec:exp1} holds that using \replytoseven{$\boldsymbol{\phi}^b$} feature vector fosters a faster convergence than using \replytoseven{$\boldsymbol{\phi}$}.

\begin{figure}[t]
    \centering
    \includegraphics[trim={1cm 0cm 2cm 1cm},clip,width=\linewidth]{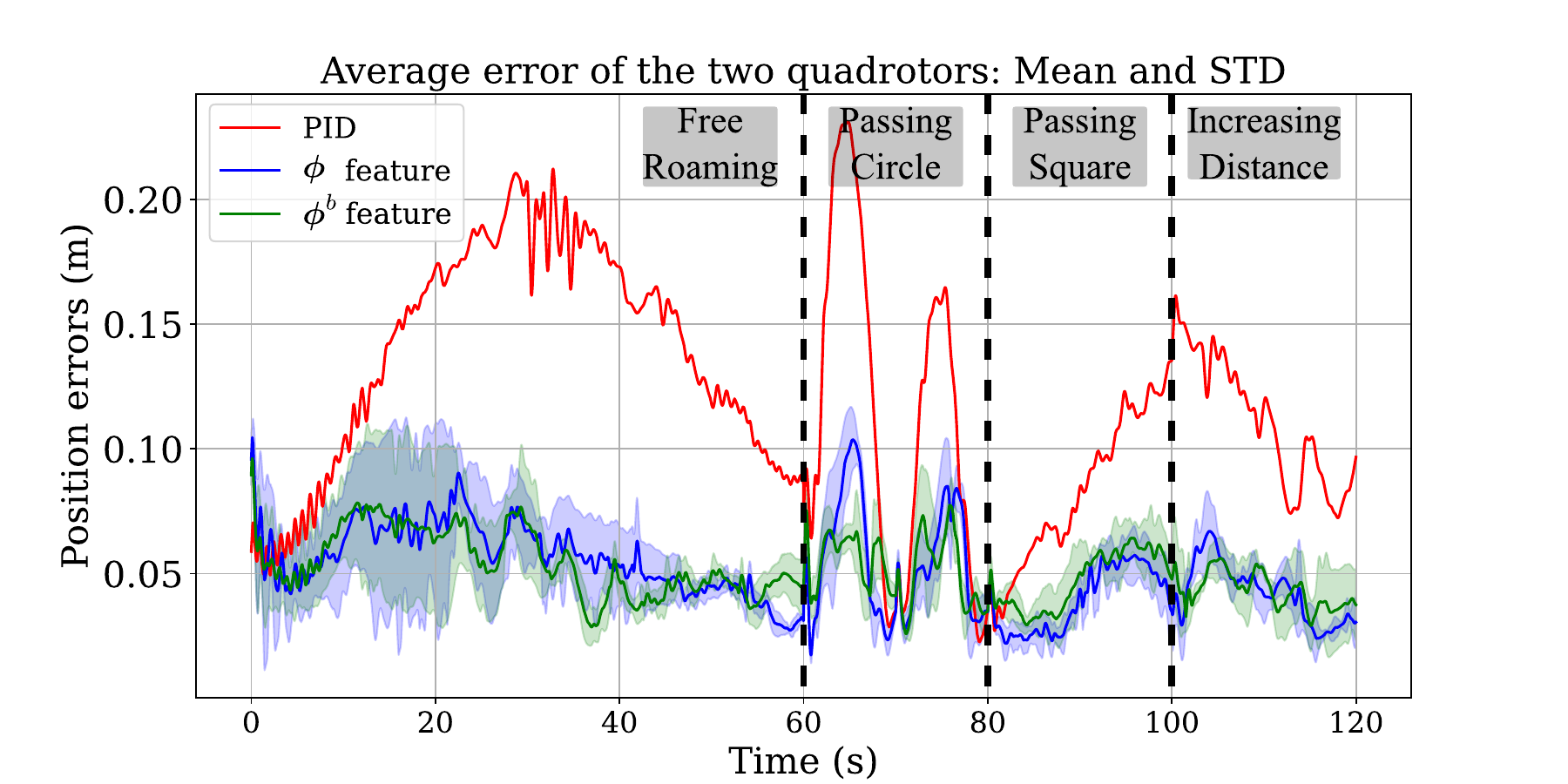}
    \caption{The mean and standard deviation of the average endpoint position errors v.s. time for passing two obstacles.}
    \label{fig:two_obs_time}
\end{figure}
\begin{figure}[t]
    \centering
    \includegraphics[trim={1cm 0cm 2cm 1cm},clip,width=\linewidth]{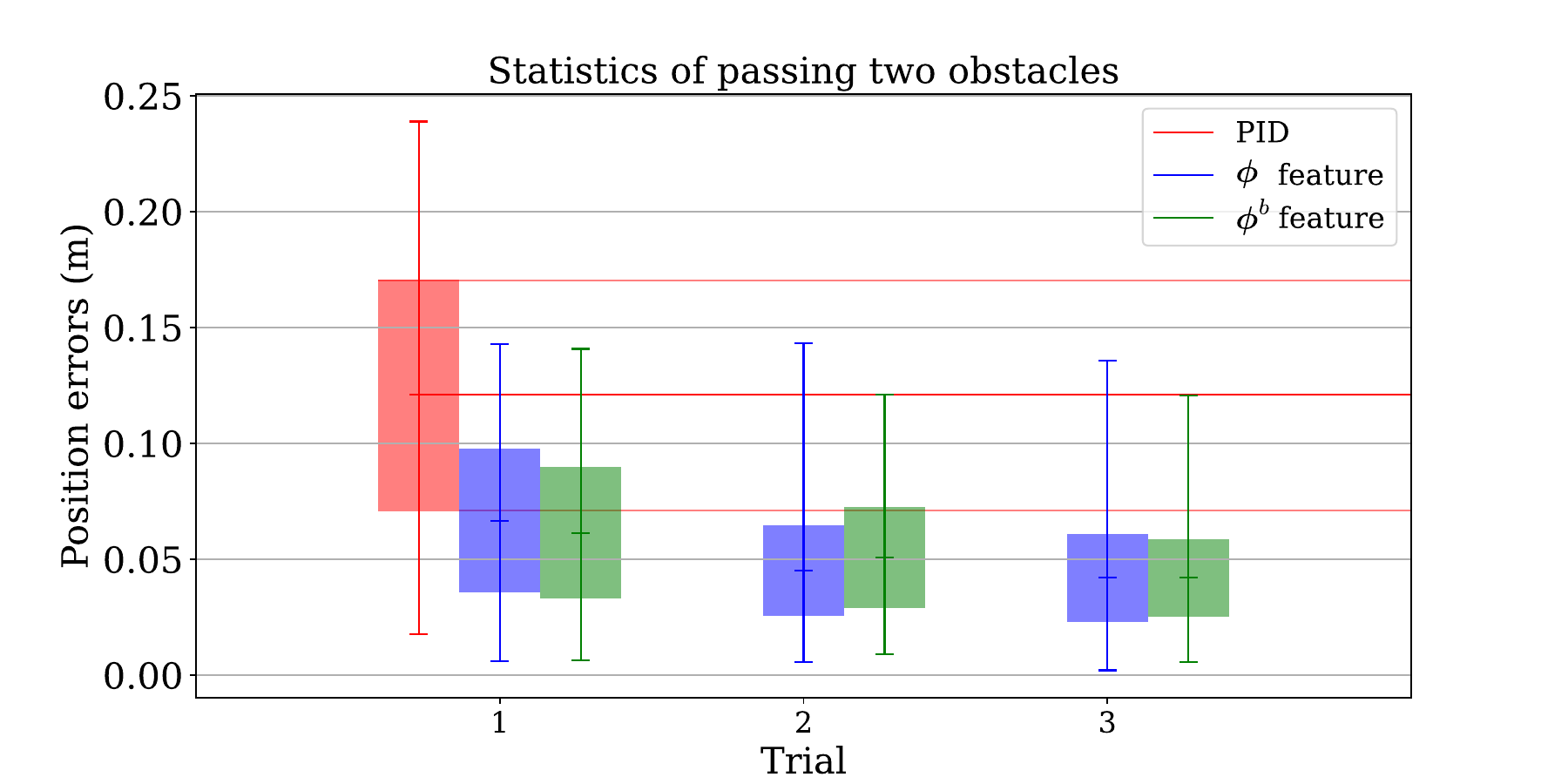}
    \caption{The mean and standard deviation of the average endpoint position errors v.s. trial for passing two obstacles.}
    \label{fig:two_obs_trial}
\end{figure}
\subsection{Complex Task: Traversing Two Perpendicular Windows}{
    % \begin{figure}[t]
    %     \centering
    %     \includegraphics[width=\linewidth]{figures/exp3.png}
    %     \caption{The quadrotors' trajectory for experiment 3.}
    %     \label{fig:exp3}
    % \end{figure}
    In the third experiment, both quadrotors first follow an obstacle-free trajectory for $60$ seconds to maintain persistence of excitation. The trajectory varies the endpoint displacement between $0.4$ m and $1.0$ m with a sinusoidal function, changes the heights of both endpoints at different frequencies with a magnitude of $0.1$ m, completes a full $2\pi$ revolution of $\mathcal{P}$, and makes the oscillatory translation of both endpoints in $x$- and $y$-axes with a magnitude of $4.0$ m and $0.2$ m, respectively. In the second $60$-second period, the two quadrotors transport the bendable strip to pass through a circular window while bending it with a fixed $\mathcal{P}$ orientation, then pass through a second square window with a varying $\mathcal{P}$ orientation. Then, the quadrotors recover the endpoint displacement to the take-off value before landing. This $120$-second experiment emulates real-world scenarios where two quadrotors transport a bendable object through a cluttered environment while constantly bending it to avoid collisions.

    The results of three trials using the adaptive controllers and the baseline PID controller are shown in Fig.~\ref{fig:two_obs_time} and~\ref{fig:two_obs_trial}. Overall, our adaptive controllers with both feature vectors converge to a mean error half that achieved by the PID controller in the first trial and continue to achieve significantly lower errors than the PID controller in the follow-up experiments. The results demonstrate the effectiveness of our adaptive controller in realistic settings. 
    \replytoall{Since the integral term only compensates for accumulated error in the world frame, it does not account for the changing orientation of the displacement vector presented in this experiment. Consequently, our adaptive controller achieves significantly higher performance facing varying force directions in the world frame than the PID baseline.}
}

\section{Conclusion}
In this paper, we presented the problem of using aerial vehicles to transport a bendable object. Due to the limited actuation capabilities of aerial vehicles and their need to maintain airborne stability, the unknown elastic forces resulting from bending the object might lead to crashes if not properly addressed. We modeled the unknown force as a parametric function of the bendable object's endpoint displacement and proposed an adaptive controller based on recursive least squares to estimate the force. By integrating the force adaptation into the trajectory tracking controller, we derived a novel controller that allows aerial vehicles to transport the bendable object while adapting to the elastic forces in real time. Through Lyapunov analysis, we proved the stability of the trajectory tracking and \replytoseven{the convergence of parameter approximation for force adaptation}. We demonstrated the effectiveness of our methods with real quadrotors and carbon-fiber strips in extensive experiments featuring different trajectories. In all tests, our adaptive controller significantly outperformed the PID controller, achieving smaller errors. In future work, we aim to extend this method to account for different object shapes, numbers of robots, and high-speed maneuvers.

\bibliographystyle{IEEEtran}
\bibliography{ref}


%\vfill


\end{document}